\pdfoutput=1

\documentclass[11pt]{article}

\usepackage[final]{acl}
\usepackage{url}

\usepackage{times}
\usepackage{latexsym}

\usepackage[T1]{fontenc}

\usepackage[utf8]{inputenc}

\usepackage{microtype}

\usepackage{inconsolata}

\usepackage{graphicx}

\usepackage{multirow}
\usepackage{subcaption}
\usepackage{pgfplots} 

\usetikzlibrary{matrix}
\usetikzlibrary{patterns}
\usepgfplotslibrary{groupplots}
\pgfplotsset{compat=newest}

\usepackage[colorinlistoftodos]{todonotes}
\usepackage{booktabs}

\usepackage{microtype}

\definecolor{tartunlp_violet}{HTML}{7268D8}
\definecolor{tartunlp_red}{HTML}{EF6650}
\definecolor{tartunlp_yellow}{HTML}{E0B12B}
\definecolor{tartunlp_black}{HTML}{282828}
\definecolor{tartunlp_green}{HTML}{4DB6AC}
\definecolor{tartunlp_blue}{HTML}{3185FF}
\definecolor{tartunlp_gray}{HTML}{9B9B9B}
\usepackage{filecontents}
\usepackage{pgfplotstable}
\usepgfplotslibrary{colorbrewer}
\usepgfplotslibrary{groupplots}

\usepackage[ukrainian,main=english]{babel}

\title{To Err Is Human, but Llamas Can Learn It Too}

\author{Agnes Luhtaru\textsuperscript{*},  \, Taido Purason\textsuperscript{*}, \, Martin Vainikko, \, Maksym Del, \, Mark Fishel\\
  Institute of Computer Science \\
  University of Tartu, Estonia \\
  \texttt{\{agnes,taido,martin,maksym,mark\}@tartunlp.ai} \\
  }

\begin{document}
\maketitle
\begin{abstract}

This study explores enhancing grammatical error correction (GEC) through artificial error generation (AEG) using large language models (LLMs). Specifically, we fine-tune Llama 2-based LLMs for error generation and find that this approach yields synthetic errors akin to human errors. Next, we train GEC Llama models with the help of these artificial errors and outperform previous state-of-the-art error correction models, with gains ranging between 0.8 and 6 F$_{0.5}$ points across all tested languages (German, Ukrainian, and Estonian). Moreover, we demonstrate that generating errors by fine-tuning smaller sequence-to-sequence models and prompting large commercial LLMs (GPT-3.5 and GPT-4) also results in synthetic errors beneficially affecting error generation models. We openly release trained models for error generation and correction and all the synthesized error datasets for the covered languages.

\end{abstract}

\renewcommand{\thefootnote}{\fnsymbol{footnote}}
\footnotetext[1]{Equal contribution.}

\section{Introduction}

The grammatical error correction (GEC) task aims to correct spelling and grammatical errors in text, making it valuable for a wide range of people. The best-performing GEC approaches currently use deep learning models \cite[and several others]{junczys-dowmunt-etal-2018-approaching, omelianchuk-etal-2020-gector, rothe-etal-2021-simple}, which are known to be data-hungry. Simultaneously, the availability of openly accessible error correction data is severely limited, even for languages typically considered high-resource in other tasks, such as German and Arabic \cite{bryant-etal-2023-grammatical}. This lack of data complicates the development of effective GEC systems for these and other even less-resourced languages. 

The scarcity of GEC data is commonly addressed through the creation of synthetic data, where errors are automatically added into correct sentences -- also called artificial error generation (AEG). In low-resource settings, the overwhelmingly most employed approach for AEG is applying random probabilistic perturbation (deletion, insertion, replacement) of words and/or characters \cite[and others]{grundkiewicz-etal-2019-neural, rothe-etal-2021-simple, naplava-straka-2019-grammatical}. Alternatives include usage of intricate hand-crafted rules and confusion sets \cite{rozovskaya-roth-2010-generating,xu-etal-2019-erroneous,kara-etal-2023-gecturk,bondarenko-etal-2023-comparative} and automatically learning to generate errors \cite{xie-etal-2018-noising, kiyono-etal-2019-empirical, stahlberg-kumar-2021-synthetic} -- also referred to as \emph{back-translation} (BT)\footnote{by analogy with the machine translation technique \cite{sennrich-etal-2016-improving}}. However, to the best of our knowledge, none of the related work on AEG uses pre-trained foundation models or applies this methodology in a low-resource setting.

This gap is precisely the focus of the present work: we are using pre-trained language models for synthetic error generation and demonstrate the simplicity and effectiveness of the approach in low-resource scenarios. We approach the task by fine-tuning open large language models (LLMs) based on Llama 2 \cite{touvron2023llama2} for error generation and correction, resulting in quality AEG data and state-of-the-art GEC models even when very limited human error data is available. Our analysis shows that the resulting errors can be categorized similarly to human errors. We also compare fine-tuning approach to prompting commercial LLMs \cite[GPT-3.5 and GPT-4:][]{openai2023gpt4} to perform AEG, as well as include other open models commonly employed for GEC and tune them for AEG: mT5 \cite{rothe-etal-2021-simple,gomez-etal-2023-low} and NLLB \cite{luhtaru2024nelb}.

Our final goal and evaluation setting is improving GEC for languages with limited GEC data. In particular, we focus on German, Ukrainian, and Estonian GEC. When pre-trained on our LLM-generated synthetic errors, the resulting GEC models achieve the best current results on the included benchmarks in all three evaluated cases, including previous state-of-the-art and 4-shot GPT-4.

We publicly release AEG and GEC models from our work and the generated data. The datasets include one million sentences for German, Ukrainian, and Estonian, each processed with three different models, as well as an additional set of 100k sentences with GPT models.

In summary, our contributions are as follows:
\begin{itemize}
    \item We show that pre-trained language models can be fine-tuned to generate high-quality synthetic errors even with limited data.
    \item We compare the influence of different models applied to AEG (LLama/GPT/mT5/NLLB) on subsequent GEC models.
    \item We achieve new state-of-the-art GEC results across all tested languages with Llama 2-based models outperforming related work as well as GPT-4.
    \item We openly release GEC and AEG models as well as AEG datasets and implementation of training and inference to facilitate future research\footnote{  \href{https://github.com/TartuNLP/gec-llm}{\tt github.com/TartuNLP/gec-llm}}.  
\end{itemize}

The paper is structured as follows. We outline related work in Section \ref{sct:relwork}, methodology experimental settings in Section~\ref{sct:experiments}, and results in Section~\ref{sct:results}. Additional questions on the same topic are discussed in Section~\ref{sct:limitations} and the paper is concluded in Section~\ref{sct:conclusion}.

\section{Related Work}
\label{sct:relwork}

The use of synthetic data is a common concept in GEC. The first effective neural method proposed by \citet{junczys-dowmunt-etal-2018-approaching} approaches GEC as low-resource Machine Translation (MT) translating from erroneous text to correct text, making it a relatively resource-heavy method encouraging synthetic data generation. Over the years, there have been different approaches to deliberately introducing errors into monolingual text, like rule-based and probabilistic methods, methods based on confusion sets and error patterns, models trained for error generation and using round-trip translation \cite{bryant-etal-2023-grammatical}.

One widely adopted approach to generating synthetic data involves the probabilistic addition of errors to monolingual corpora. This technique encompasses inserting, deleting, substituting, or moving characters or words without considering the context, as described by \citet{grundkiewicz-etal-2019-neural}, \citet{zhao-etal-2019-improving}, and \citet{rothe-etal-2021-simple}. Additionally, \citet{grundkiewicz-etal-2019-neural} introduced a "reverse speller" approach that suggests word replacements from confusion sets based on the speller's corrections. This method has been applied to several languages such as German, Czech, Russian, Ukrainian, Icelandic and Estonian \cite{naplava-straka-2019-grammatical, trinh-rozovskaya-2021-new, naplava-etal-2022-czech, gomez-etal-2023-low, ingolfsdottir-etal-2023-byte, luhtaru2024nelb}. 
As we show later, errors generated with the context-free probabilistic method differ from human errors and thus cover a much smaller number of error types, shown by significantly lower GEC recall.

Learned methods of error generation typically require more resources. Before the widespread adoption of transformers and MT, various studies explored alternative approaches for training models for error generation. For instance, \citet{felice-yuan-2014-generating} and \citet{rei-etal-2017-artificial} utilized statistical machine translation to generate errors, while \citet{xie-etal-2018-noising} and \citet{yuan-etal-2019-neural} experimented with CNNs for this purpose. Additionally, \citet{kasewa-etal-2018-wronging} investigated using RNN-based sequence-to-sequence models with attention mechanisms.

Moving towards more modern MT architectures, \citet{htut-tetreault-2019-unbearable} tested various model frameworks, including transformers, and \citet{kiyono-etal-2019-empirical} specifically employed transformer models. Both of the latter studies trained models from scratch, utilizing datasets ranging from approximately 500,000 to over a million error correction examples to train the AEG system. In contrast, our work generates up to 1 million sentences with synthetic error while using between 9k and 33k human error sentences to fine-tune the base models.

During the last few years, there has been no one error-generation method that has proved its superiority. It depends on language and available resources. For English \citet{stahlberg-kumar-2021-synthetic} train Seq2Edit models \cite{stahlberg-kumar-2020-seq2edits} from scratch for learning to create diverse sets of errors. As mentioned in the beginning, synthetic probabilistic errors have found wide use for different languages. For instance, \citet{ingolfsdottir-etal-2023-byte} combine probabilistic character/word permutations with a rule-based approach for Icelandic and \citet{kara-etal-2023-gecturk} curate special rules for generating Turkish data. 

To the best of our knowledge, no works focus specifically on error generation by LLMs; however, several studies have evaluated the performance of commercial LLMs in this task. \citet{fang2023chatgpthighlyfluentgrammatical} found that while GPT-3.5 performs significantly worse than other systems in terms of precision, it excels in recall. Similar results were reported by \citet{wu2023chatgptgrammarlyevaluatingchatgpt}, who observed that GPT models tend to overcorrect rather than undercorrect errors. This finding is also supported by \citet{coyne2023analyzingperformancegpt35gpt4}, who noted that GPT models are particularly strong at making fluency edits. While there are few studies that use or evaluate open-source LLMs, \citet{zhang2023multitaskinstructiontuningllama} explore the use of the LLaMA model \cite{touvron2023llamaopenefficientfoundation} for writing assistance.

Next, we present the key methodological details of our work.

\section{Methodology and Experiments}
\label{sct:experiments}

The primary target of our work is to apply generative language models to AEG via fine-tuning. Additionally, we experiment with prompting LLMs to perform the same task and include two seq2seq models that are fine-tuned to do the same.

The efficiency of proposed AEG solutions is evaluated by using them to improve GEC. Thus, we also fine-tune generative LLMs to perform the GEC task and compare the results to prompting-based GEC results and related work. The general pipeline of our approach is straight-forward:

\begin{enumerate}
\item[1:] Fine-tune an LLM to generate errors using human error data, with correct sentences as input and sentences with errors as output. 
\item[2:] Apply that AEG LLM to correct sentences in order to add a synthetically erroneous counterpart.
\item[3:] Fine-tune an LLM on that synthetic dataset to correct grammatical errors. Equivalent to Step 1, with the sentence pair direction reversed.
\item[4:] Continue fine-tuning GEC LLM on the smaller dataset with human errors.
\item[5:] Apply the models to the erroneous sentences of the benchmark test sets and evaluate the results
\end{enumerate}

Next, we describe the technical details of our implementation and the experimental setup.

\subsection{Data}
We use two distinct types of data in our work. Firstly, we rely on datasets containing examples of corrections to train our error generation systems and correction models. Secondly, we incorporate monolingual data to create synthetic datasets by introducing errors. See an overview of used data in Table \ref{tab:data}.

We use the language learners' corpus from the University of Tartu (UT-L2 GEC) \cite{rummo2017error} for gold data in Estonian. In Ukrainian, we use the UA-GEC corpus \cite{syvokon-etal-2023-ua} used in the UNLP 2023 Shared Task on Grammatical Error Correction for Ukrainian \cite{syvokon-romanyshyn-2023-unlp}, using the GEC+Fluency data for training. For German, we rely on the widely used Falko-Merlin (FM) corpus \cite{boyd-2018-using}.

For monolingual Estonian data, we employ the Estonian National Corpus 2021 \cite{enc21}. We randomly sample equal sets from the latest Wikipedia, Web, and Fiction subsets and shuffle these together. For Ukrainian and German, we use the CC-100 dataset \cite{conneau-etal-2020-unsupervised, wenzek-etal-2020-ccnet}. Depending on the experiments, we sample the required number of sentences from the larger corpora (i.e., one million or 100 thousand sentences or a set equal to gold corpora sizes).

\begin{table}
    \centering
    \begin{tabular}{lccc}
    Corpus &  Language & Train & Test \\ 
    \toprule

       UT-L2 GEC  & ET & 8,935  & - \\
       EstGEC-L2 & ET & - & 2,029 \\
       UA-GEC  & UK & 31,038 &  1,271 \\
        FM & DE & 19,237  &  2,337\\
        \midrule
        ENC 2021 & ET & 1M/100k & - \\
        CC-100 & UK/DE & 1M/100k & - \\
    \bottomrule
    \end{tabular}
    \caption{Data used for training and testing.}
    \label{tab:data}
\end{table}

\subsection{Models and Training}
\textbf{Llama-2-based models}. We fine-tune models that have been enhanced with bilingual capabilities using continued pre-training from Llama-2-7B \citep{touvron2023llama2}. For Estonian, we use Llammas-base\footnote{\href{https://huggingface.co/tartuNLP/Llammas-base}{\tt huggingface.co/tartuNLP/Llammas-base}} \cite{kuulmets-etal-2024-teaching}, and for German, LeoLM\footnote{\href{https://huggingface.co/LeoLM/leo-hessianai-7b}{\tt huggingface.co/LeoLM/leo-hessianai-7b}}. For Ukrainian, we apply continued pre-training to replicate the conditions of Estonian LLM by training with 5B tokens from CulturaX \citep{nguyen2023culturax} with 25\% of the documents being in English and the rest in Ukrainian following \citet{kuulmets-etal-2024-teaching}. For GEC and AEG fine-tuning, we formatted the training data with a prompt (see Table~\ref{tab:llama-instruction-format}~and~\ref{tab:llama-instruction-format-aeg}) loosely based on Alpaca \citep{alpaca}. During fine-tuning, the loss is calculated on the tokens of the correct sentence.
Fine-tuning details (including hyperparameters) are discussed in Appendix~\ref{sec:appendix-training-llama}.

\textbf{Other models} we use are NLLB \cite{nllbteam2022language} and mT5 \cite{xue-etal-2021-mt5}. Specifically, we use the NLLB-200-1.3B-Distilled and mt5-large (1.2B parameter) models for our experiments and train NLLB models using Fairseq \cite{ott-etal-2019-fairseq} and mT5 with HuggingFace Transformers \cite{wolf2020huggingfaces}. When training in two stages, first with synthetic data and later with human errors, we keep the state of the learning rate scheduler, following the fine-tuning approach rather than retraining as defined by \citet{grundkiewicz-etal-2019-neural}.  See Appendices~\ref{sec:appendix-training-nllb} and \ref{sec:appendix-training-mt5} for further details.

\subsection{Generation}
\textbf{Fine-tuned models}. We use sampling instead of beam search to generate the synthetic errors and sample from the top 50 predictions with a temperature of $1.0$. During error correction, beam search with a beam size of 4 is used without sampling as regularly.

\textbf{Prompt engineering}. We perform iterative prompt engineering, analyzing intermediate qualitative results and updating the prompt. For instance, we initially started with a simple 2-shot prompt (temperature = 0.1) asking GPT-3.5 to add grammatical and spelling mistakes into the input text but noticed that some error types were missing. We then improved the prompt by specifying the missing error types, adding two more examples, and upping the temperature. Our final prompt uses four examples and a model temperature of 1.0. See Appendix \ref{sec:appendix-prompts} for the prompts. We randomly pick the examples from each language's train set for few-shot prompting. When comparing the prompting between GPT-4-Turbo and GPT-3.5-Turbo, we use an identical random set of examples to ensure comparability.

Finally, we converged on using GPT-3.5-turbo for more massive error generation (100,000 sentence pairs per language). The motivation for that is partially financial (as GPT-4/GPT-4-turbo are several times more expensive) as well as performance-driven (see Figure~\ref{fig-qual} and description for details).

We apply simple post-processing to the resulting set because, in some cases, parts from the prompt are duplicated in the output. If the model didn't generate a response due to safety model activation or the response was too short or too long compared to the input sentence, we replaced the output with the source text (equivalent to adding no errors).

The precise model versions we prompt are \verb|gpt-4-1106-preview| for GPT-4-Turbo (using the OpenAI API) and \verb|gpt-3.5-turbo| (GPT-3.5-Turbo) and \verb|gpt-4| (GPT-4) (using Azure OpenAI API, version 0613 for both).

\textbf{Probabilistic errors}. We generate rule-based synthetic errors as done in prior work \cite{grundkiewicz-etal-2019-neural, naplava-straka-2019-grammatical, gomez-etal-2023-low, luhtaru2024nelb} using the same method and also employing the Aspell speller\footnote{\href{http://aspell.net/}{\tt aspell.net}} for replacing subwords.

\subsection{Automatic Evaluation of Models}

We evaluate the performance of our GEC models using test sets and evaluation metrics consistent with those employed in previous works (see datasets in Table \ref{tab:data}).

For Estonian, we evaluate our models using the Estonian learner language corpus (EstGEC-L2)\footnote{\href{https://github.com/tlu-dt-nlp/EstGEC-L2-Corpus/}{\tt github.com/tlu-dt-nlp/EstGEC-L2-Corpus}}, alongside a modified version of the MaxMatch scorer\footnote{\href{https://github.com/TartuNLP/estgec/tree/main/M2_scorer_est}{\tt github.com/TartuNLP/estgec}}, following \citet{luhtaru2024nelb}. The Estonian scorer also outputs recall per error category, accounting for both other errors within the word order error scope and not accounting for these. We report the ones that do consider other errors separately.  For Ukrainian, our evaluation methodology aligns with that of the UNLP 2023 Shared Task \cite{syvokon-romanyshyn-2023-unlp}, utilizing the CodaLab platform for submissions to a closed test set that uses the ERRANT scorer for evaluation \cite{bryant-etal-2017-automatic}. We follow the GEC+Fluency track setting since it encompasses a wider range of challenging errors. For German, we use the test set from the Falko-Merlin (FM) corpus \cite{boyd-2018-using} that several works have reported their scores on and the original MaxMatch scorer \cite{dahlmeier-ng-2012-better}.

\subsection{Human Evaluation of Generated Data}

In addition to evaluating the quality of our data in terms of its usefulness for training better models, we perform a detailed evaluation of generated data in Estonian. We apply the same annotation scheme \citet{annotations-est} used for annotating test and development sets to artificially generated sentences. This comparison allows us to assess the error distribution between the training data and generated data and to see whether the errors can be categorized into the same classes.

We select 100 random sentences from sets generated by Llama-based models, GPT-3.5-Turbo and GPT-4-Turbo\footnote{We also considered annotating probabilistic denoising errors, but these contained very few edits that could be categorized based on the annotation scheme.}, for annotation and also annotate 100 sentences from the training set. We add labels for problematic errors generated by the model, such as hallucinations and truncation of words important for understanding the meaning of sentence (HALL), synonym swaps (SYN), optional edits (O), corrections of mistakes in original sentences (INACC), and transformations that make the original word unrecognizable (UNREC).

\section{Results}
\label{sct:results}

\begin{table*}[h]
\centering
\begin{tabular}{lccccccccc}
\toprule
\multirow{2}{*}{Method }& \multicolumn{3}{c}{Estonian} & \multicolumn{3}{c}{Ukrainian} & \multicolumn{3}{c}{German} \\
\cmidrule(lr){2-4} \cmidrule(lr){5-7} \cmidrule(lr){8-10}
 & P & R & F$_{0.5}$ & P & R & F$_{0.5}$&  P & R & F$_{0.5}$\\
\midrule
GPT-4-Turbo (4-shot) & 70.86 & 57.35 & 67.67 & 39.62 & 42.13 & 40.1 & 64.15 & 69.34 & 65.12 \\
GPT-4 (4-shot) & 70.04 &\textbf{ 59.03} & 67.52 &36.25 & 37.77 & 36.54 & 65.22 & \textbf{69.75 }& 66.08 \\
\midrule
Old SOTA & 71.27 & 55.38 & 67.40 & 79.13 & 43.87 & 68.17 & 78.5 & 68.4 & 76.3 \\
\midrule
Llama + gold & 71.52 & 55.23 & 67.54 & 79.98 & 51.76 & 72.12 & 76.86 & 65.60 & 74.31 \\
Llama + prob + gold & 72.59 & 54.72 & 68.14 & 80.37 & 53.19 & 72.92 & 78.22 & 67.65 & 75.85 \\
Llama + BT + gold & \textbf{73.85}\textsuperscript{†} & 57.83\textsuperscript{†} & \textbf{69.97}\textsuperscript{†} & \textbf{82.03} & \textbf{53.41} &\textbf{ 74.09} &  \textbf{79.08}\textsuperscript{†} & 68.66 & \textbf{76.75}\textsuperscript{†} \\
\bottomrule
\end{tabular}
\caption{Comparison of Llama 2-based models (denoted as Llama) after extended pre-training and GEC fine-tuning: Models without synthetic data (LLM + gold) versus models with synthetic data generated with a probabilistic reverse-speller method (LLM + 1M prob + gold) and back-translation style learned synthetic data (LLM + 1M BT + gold). State-of-the-art benchmarks include \citet{luhtaru2024nelb} for Estonian (NLLB-200-1.3B-Distilled with mixed synthetic and translation data training), \citet{bondarenko-etal-2023-comparative} for Ukrainian (mBART-based model with synthetic data), and \citet{fang-etal-2023-improving} for German (multimodal mixture-of-experts based on mT5). † - significant improvement compared to Llama + 1M prob + gold according to paired bootstrap resampling significance test \cite{koehn-2004-statistical} with 10,000 samples and $p=0.05$. Significance testing was not possible for Ukrainian due to closed test set. }
\label{tab:combined_results}
\end{table*}

\begin{table}
    \centering
    \begin{tabular}{lccc}
        \toprule
        Lang/Model & Llama & NLLB & mT5 \\
        \midrule
        ET (AEG only) & 65.30 & 65.34 & 59.40 \\
        ET (AEG + gold) & 69.97 & 69.73 & 68.57 \\
        \midrule
        UK (AEG only) & 28.39 & 27.04 & 16.79 \\
        UK (AEG + gold) & 74.09 & 72.30 & 72.51 \\
        \midrule
        DE (AEG only) & 71.29 & 69.13 & 54.96 \\
        DE (AEG + gold) & 76.75 & 76.28 & 74.77 \\
        \bottomrule
    \end{tabular}
    \caption{F$_{0.5}$-scores for Llama-based models fine-tuned with 1M sentences generated with different AEG models and then further fine-tuned with gold GEC data. The errors are generated with 7B Llama-2-based models, 1.3B NLLB model and 1.2B mT5 model.}
    \label{tab:small-models-f}
\end{table}

In this section, we evaluate the performance of Llama-based models for GEC and AEG tasks. We then compare the AEG effectiveness between NLLB and mT5 models against Llama-based models to see if smaller, more efficient models can generate quality data. Separately, we assess AEG through prompting with GPT-3.5-turbo versus Llama models with trained error generation. Finally, we examine the quality of generated errors against human data and probabilistic reverse-speller errors and compare the error type distributions for Estonian.

\begin{table*}
\centering
\begin{tabular}{lcccccc}
\toprule
 & \multicolumn{3}{c}{Prompting} & \multicolumn{3}{c}{Fine-tuning} \\
\multirow{2}{*}{Lang/Model} & \multicolumn{3}{c}{GPT-3.5-turbo (100k)} & \multicolumn{3}{c}{Llama (100k)} \\
\cmidrule(lr){2-4} \cmidrule(lr){5-7}
& P & R & F$_{0.5}$ & P & R & F$_{0.5}$ \\
\midrule
ET (AEG only) & 71.72 & 44.20 & 63.78 & 67.57 & 50.89 & 63.41 \\
ET (AEG + gold) & 71.11 & 56.56 & 67.63 & 71.51 & 56.51 & 67.91 \\
\midrule
UK (AEG only) & 28.61 & 22.16 & 27.04 & 40.00 & 19.87 & 33.26 \\
UK (AEG + gold) & 80.82 & 51.33 & 72.49 & 80.89 & 50.31 & 72.12 \\
\midrule
DE (AEG only) & 70.55 & 49.61 & 65.05 &  70.07 & 59.11 & 67.56  \\
DE (AEG + gold) & 78.06 & 67.06 & 75.58 & 78.80 & 67.52 & 76.25 \\
\bottomrule
\end{tabular}
\caption{Scores of Llama-based models fine-tuned with 100k sentences generated by Llama-based model fine-tuned for error generation and GPT-3.5-model prompted to add errors.}
\label{tab:prompting-vs-fine-tuning}
\end{table*}

\subsection{Artificial Error Generation and Correction with Llama}

We compare LLama-based LLM fine-tuning error corrections across three configurations: (1) the baseline approach of training exclusively on human error GEC data, (2) the established related work approach of training on probabilistic reverse-speller AEG data and then continuing training with human error GEC data, and (3) our approach of training on back-translation style AEG data produced by fine-tuned Llama-based models first, followed by fine-tuning on human data. 

The resulting scores are compared in Table~\ref{tab:combined_results}, along with previous state-of-the-art (SOTA) scores and results of GEC via 4-shot prompting of GPT-4/GPT-4-turbo. Results show that Llama-based models, further enhanced through continued pre-training, exhibit strong correction capabilities across languages in our study. Even without synthetic data, these models outperform current SOTA methods in Estonian and Ukrainian error correction, and are not too far behind in German, trailing the best score by two points. When comparing our 7B Llama model to others, there are significant differences in model sizes and data usage that need to be considered for a fair evaluation. Our 7B Llama model is substantially larger than the NLLB-200-1.3B-Distilled model used for Estonian \cite{luhtaru2024nelb} and the mBART model used for Ukrainian \cite{bondarenko-etal-2023-comparative}. However, it is smaller than the 13B gT5-xxl model, which represents the current state-of-the-art for German text-only data \cite{rothe-etal-2021-simple}, while it is larger than the multimodal German model incorporating both text and speech data \cite{fang-etal-2023-improving}. In terms of synthetic data usage, our model is trained with one million sentences, which contrasts with the six million sentences per language used by \citet{luhtaru2024nelb} in their multilingual training approach, and the smaller, more carefully crafted synthetic datasets used by \citet{bondarenko-etal-2023-comparative}. Notably, all these models rely on the same human-labeled data, ensuring consistency in that aspect.

Incorporating synthetic data as a preliminary step to fine-tuning significantly enhances performance across all languages and synthetic data types. Notably, our back-translation style synthetic data consistently delivers superior precision and recall compared to the probabilistic reverse-speller method. This approach results in a 2-2.4 point increase in the F$_{0.5}$ score relative to solely using gold data for fine-tuning. Conversely, the gains from using probabilistic reverse-speller data are more modest, ranging from 0.6 to 1.5 points, highlighting the enhanced utility of our learned AEG errors.

Our systems consistently outperform GPT-4 models in terms of precision across all languages studied. However, GPT-4 models exhibit higher recall rates for Estonian and German. This discrepancy indicates that while our systems are more accurate in identifying correct instances, GPT-4 models better retrieve a broader range of relevant errors in these languages.GPT-4’s performance on the Ukrainian test set is significantly lower compared to other methods and languages, likely due to the distinctive features of the dataset. Unlike the Estonian and German datasets, the Ukrainian set contains a higher proportion of punctuation errors (43\%) and has a two times smaller error rate than German (8.2 vs 16.8) \cite{syvokon-etal-2023-ua}. Since recent studies show that GPT models struggle with punctuation errors and tend to make more extensive changes to sentences \cite{katinskaia-yangarber-2024-gpt}, this likely explains the variation in performance.

\subsection{Artificial Error Generation with Smaller Models}

Since error generation with 7B Llama-based models can be costly and time-consuming and many other architectures have proved useful for correction, we also explore smaller models for AEG: the 1.3B NLLB model and 1.2B mT5-large. The goal here is to see if these can also produce useful errors. 

Table~\ref{tab:small-models-f} shows the results of the analysis. Both models can learn valuable information that improves performance beyond what is achieved with fine-tuning on gold data alone. Notably, errors generated by the NLLB model are particularly effective, delivering results close to those achieved by LLM-generated errors in Estonian and German, almost matching the performance of LLama-based models. However, for Ukrainian, NLLB-generated errors fall behind probabilistic reverse-speller errors. This is likely because the dataset contains many special punctuation characters that get normalized during preprocessing (see more in Appendix~\ref{sec:nllb-zs-correction}). 

The mT5 models, in contrast, appear less adept at error generation. The errors produced by mT5 lag behind those from probabilistic reverse speller for Ukrainian and German and offer only a minimal improvement for Estonian.

We can also see that the scores before gold fine-tuning highlight that Ukrainian scores are notably low across all methods. However, these scores recover well after fine-tuning, suggesting the synthetic data may not align well with the text domain or error types specific to the Ukrainian language. This can be related to the unique characteristics of the Ukrainian dataset, which also causes GPT-4 to struggle with the GEC task. Estonian and German models show higher scores for models trained with just AEG data and improve less drastically with fine-tuning.

\subsection{Artificial Error Generation with Prompting}

To assess the capability of generating errors without additional LLM training, we utilize advanced commercial models, specifically exploring the efficiency of error generation through prompting GPT-3.5-turbo with datasets comprising 100,000 sentences. We later also explore the effectiveness of GPT-4-Turbo in a more limited setting (see Section~\ref{sec:qual}).

The generation cost depends on the sum of input and completion tokens. Ukrainian, our most expensive language, had the highest number of tokens per 100,000 sentences: 98 million input and 12 million completion tokens. The cost for input tokens with GPT-3.5-Turbo in USD is \$147, and for completion tokens, it is \$25 -- in total, \$172 for generating 100,000 Ukrainian sentences. In comparison, the costs with GPT-4-Turbo would have been \$983 and \$370, respectively\footnote{\url{https://openai.com/pricing}}. 

Table \ref{tab:prompting-vs-fine-tuning} shows the results of continued pre-training Llama-based models on the same amount of sentences (100,000) with synthetic errors from prompting or fine-tuning. In terms of error correction quality after gold fine-tuning, employing GPT-3.5-turbo for prompting and fine-tuning Llama-2-based models are both viable strategies for AEG, as they lead to very close F$_{0.5}$ scores in all three languages (with a slight difference in favor of fine-tuning errors for German: 75.58 vs 76.25).

Analyzing the performance before gold fine-tuning reveals distinct differences between the two methods. For Estonian and German, recall rates are significantly higher with fine-tuning than prompting, though precision is slightly compromised. Conversely, Ukrainian exhibits the reverse pattern. However, it's important to note that any disparities observed before gold fine-tuning are greatly diminished after training on actual error correction examples. The most considerable remaining difference is under 0.7 points for German, with smaller discrepancies for Estonian and Ukrainian.

When comparing LLama model scores for 100k to the ones with only gold tuning (see Table~\ref{tab:combined_results}), we can see that although scores increase more modestly, only 100k examples of synthetic data increase the scores more for German (almost 2 F$_{0.5}$-score points), a bit for Estonian (around 0.4 points) and stay the same for Ukrainian with higher precision and lower recall. The scores for models trained with 100k sentences are mostly lower than those trained with 1M reverse-speller errors, which indicates that the data quantity jump from 100,000 to 1M plays a significant role.

\pgfplotstableread{
71.52	55.23	67.54
69.02	45.85	62.68
63.25	47.48	59.31
67.52	46.55	61.94
70.49	18.74	45.41
}\datatableEST

\pgfplotstableread{
79.98	51.76	72.12
34.15	22.16	30.81
34.65	22.44	31.25
45.83	19.97	36.4
32.38	10.26	22.63
}\datatableUKR

\pgfplotstableread{
76.86	65.6	74.31
70.65	47.02	64.2
67.38	46	61.65
70.64	53.43	66.37
64.02	17.73	42.05
}\datatableDEU
    
\begin{figure*}[h!]
    \centering

    \begin{tikzpicture}
    \begin{groupplot}[
      group style={
        group size=3 by 1,
        horizontal sep=1.25cm,
        group name=plots
      },
        width  = 0.33\textwidth,
        height = 4.5cm,
        major x tick style = transparent,
        x axis line style = {opacity=0},
        y axis line style = {opacity=0},
        axis y line = left,
        ylabel near ticks,
        ymajorgrids = true,
        ytick style={draw=none},
        xticklabels = {
            P,
            R,
            F\textsubscript{0.5}
        },
        xtick = data,
        title style = {font=\small},
        /pgf/bar width={3.5pt},  
        ybar,
        xticklabel style = {align=center, font=\small},
        yticklabel style = {font=\tiny},
        enlarge x limits=0.25,
        ymin=0, ymax=85,
        legend cell align=left,
        label style={font=\tiny},
    ]
    
        \pgfplotstabletranspose\datatransposedEST{\datatableEST} 
        \pgfplotstabletranspose\datatransposedDEU{\datatableDEU} 
        \pgfplotstabletranspose\datatransposedUKR{\datatableUKR} 
        
        \nextgroupplot[
      legend entries={Gold, GPT-3.5, GPT-4-turbo, LLM, Prob},
      legend to name=thelegend,
      title={Estonian}]
            \addplot[style={tartunlp_yellow, fill=tartunlp_yellow,mark=none}]
                 table[y index=1]{\datatransposedEST};
    
            \addplot[style={tartunlp_red,fill=tartunlp_red,mark=none}]
                 table[y index=2]{\datatransposedEST};
    
            \addplot[style={tartunlp_violet,fill=tartunlp_violet,mark=none}]
                 table[y index=3]{\datatransposedEST};
    
            \addplot[style={tartunlp_green,fill=tartunlp_green,mark=none}]
                 table[y index=4]{\datatransposedEST};
    
            \addplot[style={tartunlp_gray,fill=tartunlp_gray,mark=none}]
                 table[y index=5]{\datatransposedEST};
    
        \nextgroupplot[
      legend entries={Gold, GPT-3.5, GPT-4-turbo, LLM, Prob},
      legend to name=thelegend,
      title={German}]
            \addplot[style={tartunlp_yellow, fill=tartunlp_yellow,mark=none}]
                 table[y index=1]{\datatransposedDEU};
    
            \addplot[style={tartunlp_red,fill=tartunlp_red,mark=none}]
                 table[y index=2]{\datatransposedDEU};
    
            \addplot[style={tartunlp_violet,fill=tartunlp_violet,mark=none}]
                 table[y index=3]{\datatransposedDEU};
    
            \addplot[style={tartunlp_green,fill=tartunlp_green,mark=none}]
                 table[y index=4]{\datatransposedDEU};
    
            \addplot[style={tartunlp_gray,fill=tartunlp_gray,mark=none}]
                 table[y index=5]{\datatransposedDEU};

        \nextgroupplot[
      legend entries={Gold, GPT-3.5-turbo, GPT-4-turbo, Llama, Prob},
      legend to name=thelegend,
      title={Ukrainian},
      legend style={
                    font=\tiny,
                    column sep=0.5ex,
                    draw=none,
                    legend columns = -1,
            },
        legend style={/tikz/every even column/.append style={column sep=0.5cm}}]
      ]
            \addplot[style={tartunlp_yellow, fill=tartunlp_yellow,mark=none}]
                 table[y index=1]{\datatransposedUKR};
    
            \addplot[style={tartunlp_red,fill=tartunlp_red,mark=none}]
                 table[y index=2]{\datatransposedUKR};
    
            \addplot[style={tartunlp_violet,fill=tartunlp_violet,mark=none}]
                 table[y index=3]{\datatransposedUKR};
    
            \addplot[style={tartunlp_green,fill=tartunlp_green,mark=none}]
                 table[y index=4]{\datatransposedUKR};
    
            \addplot[style={tartunlp_gray,fill=tartunlp_gray,mark=none}]
                 table[y index=5]{\datatransposedUKR};
            

            \legend{Gold, GPT-3.5, GPT-4, Llama, Prob}
            \coordinate (c2) at (rel axis cs:1,1);
        \end{groupplot}
        \coordinate (c3) at ($(rel axis cs:0,1)!.5!(c2)$);

        \node[below] at (c3 |- current bounding box.south)
          {\pgfplotslegendfromname{thelegend}};

    \end{tikzpicture}
    \caption{Quality of generated errors compared to gold and probabilistic, as shown by GEC results of tuning Llama-based models on same-sized synthetic or human (gold) error sets. GPT-3.5-turbo and GPT-4-turbo errors are generated via prompting, Llama stands for Llama 2-based model fine-tuned on the AEG task.}
    \label{fig-qual}
\end{figure*}
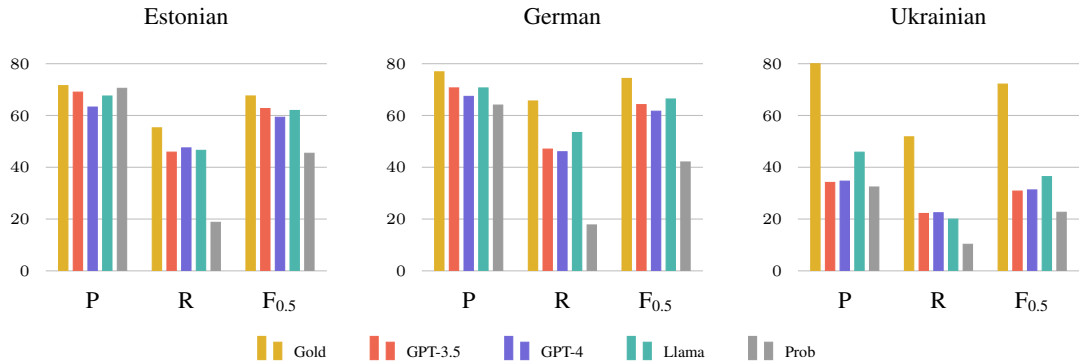
\pgfplotstableread{
70	45	82	47	28	81	66	42	62
66	23	81	41	23	57	69	30	52
71  25  81  44  21  48  75  27  59
66	32	75	42	29	55	66	34	58
31	1	70	24	10	26	42	26	16
}\datatableErr

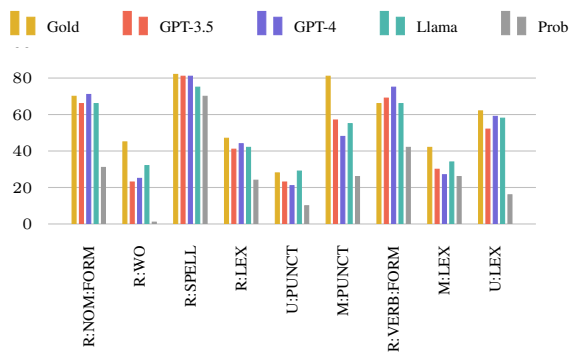
\begin{figure}
\begin{tikzpicture}
	\pgfplotstabletranspose\datatransposed{\datatableErr} 
    \begin{axis}[
        width  = 0.5\textwidth,
        height = 4cm,
        major x tick style = transparent,
        x axis line style = {opacity=0},
        y axis line style = {opacity=0},
        axis y line = left,
        ylabel near ticks,
        ymajorgrids = true,
        ytick style={draw=none},
        xticklabels = {
            R:NOM:FORM,
            R:WO,
            R:SPELL,
            R:LEX,
            U:PUNCT,
            M:PUNCT,
            R:VERB:FORM,
            M:LEX,
            U:LEX,
        },
        xtick = data,
        /pgf/bar width={1.5pt},  
        ybar=1.25pt,
        xticklabel style = {align=center, font=\tiny, rotate=90},
        yticklabel style = {font=\tiny},
        enlarge x limits=0.1,
        ymin=0, ymax=100,
        legend cell align=left,
        label style={font=\small },
        legend style={
                at={(0.5,1.2)}, anchor=north,
                font=\tiny,
                column sep=0.5ex,
                draw=none,
                legend columns = -1
        },
        legend style={/tikz/every even column/.append style={column sep=0.5cm}}]

    ]
        \addplot[style={tartunlp_yellow, fill=tartunlp_yellow,mark=none}]
             table[y index=1]{\datatransposed};

        \addplot[style={tartunlp_red,fill=tartunlp_red,mark=none}]
             table[y index=2]{\datatransposed};

        \addplot[style={tartunlp_violet,fill=tartunlp_violet,mark=none}]
             table[y index=3]{\datatransposed};

        \addplot[style={tartunlp_green,fill=tartunlp_green,mark=none}]
             table[y index=4]{\datatransposed};

        \addplot[style={tartunlp_gray,fill=tartunlp_gray,mark=none}]
             table[y index=5]{\datatransposed};


        \legend{Gold, GPT-3.5, GPT-4, Llama, Prob}
    \end{axis}
\end{tikzpicture}
    \caption{Recall scores for most frequent categories in Estonian EstGEC-L2 test set. The first letter corresponds to the operation type (R - replaced, M - missing, U - unnecessary).}
    \label{fig:err-cat}
\end{figure}

\subsection{Quality Compared to Human Data}
\label{sec:qual}

Finally, we run a direct comparison between human errors and artificial ones. To do so we train models using the same number of sentences as the respective human error set sizes: 19k sentence pairs for German, 33k for Ukrainian, and 9k sentence pairs for Estonian. We include comparing these models to ones based on one million probabilistic sentences.

Our findings indicate that the precision of all synthetic data closely matches that of high-quality (gold) data in both Estonian and German, as illustrated in Figure \ref{fig-qual}. A notable distinction, however, is observed in recall rates. For Estonian and German, the recall for errors generated by LLMs is more comparable to human-generated (gold) data than errors produced through probabilistic methods. 

Ukrainian scores with synthetic data are substantially worse than gold data, regardless of the AEG method. Still, recall for LLM-generated errors is significantly higher than for simple probabilistic errors. This might be due to a larger mismatch in the text domain or error frequency. The dataset is heavily composed of punctuation errors, which could be more challenging for LLMs to generate, as they have been shown to struggle with correcting such errors \cite{katinskaia-yangarber-2024-gpt}.

Comparing GPT-3.5-Turbo with GPT-4-Turbo, we find similar performance overall. However, for Estonian, GPT-4-Turbo exhibits higher recall but lower precision. For German, GPT-4-Turbo shows reductions in both precision and recall. Performance is nearly identical for Ukrainian between the two models. Overall, the F$_{0.5}$ scores of GPT-4-Turbo are slightly lower for Estonian and German and around the same for Ukrainian compared to GPT-3.5-Turbo.

When analyzing the recall for various error categories in Estonian, it is evident that our models trained with AEG data particularly face challenges in inserting missing punctuation marks and correcting errors related to word order, as depicted in Figure \ref{fig:err-cat}. Errors generated probabilistically excel in identifying spelling mistakes and can correct certain errors in noun and verb forms. However, they generally perform poorly in addressing issues beyond spelling errors. 

\subsection{Evaluation of Generated Errors: Case Study with Estonian}
\label{results-human-eval}
We labeled 100 LLM-generated sentences from different sets to determine if the errors made by models are similar to those in the training corpus. 

Based on the annotations, we can categorize a large proportion of the changes according to the annotation scheme, but there is still a considerable amount of problematic edits (\~25-45\%) (see Figure \ref{fig:human-eval} and Table \ref{tab:problematic} in Appendix \ref{sec:human-eval}). The human evaluation also indicates that the models differ in their error rates. GPT models generate fewer problematic errors overall, but the error category distribution seems more similar to human data with Llama-based models. This is likely due to a fine-tuning approach instead of prompting.

As mentioned in the last section, compared to human data, all models trained with generated data, correct far fewer word order and missing punctuation errors, and lexical changes are not well corrected either. These results can be partially explained by examining the different error types in generated data, where the same types are not as well represented as in human data. Most problematic edits involve generating lexical errors, which often were synonymous or changed the original meaning of the sentence, which could explain the poor performance in correcting lexical errors.  On the other hand, verb or nominal form and spelling errors were better or almost as well corrected as by a model trained with gold data, and the data contained more errors in these categories. This shows that correction recall is closely tied to the error types present in the training data, and the data generated with our approach generates realistic error types that help correction in these categories.

\pgfplotstableread{
110 49 27 53 4 21 27 19 27 4
53 13 48 15 7 7 14 6 7 102
163 18 76 13 2 5 51 3 16 130
89 34 18 16 12 7 18 8 16 212
}\datatableErr

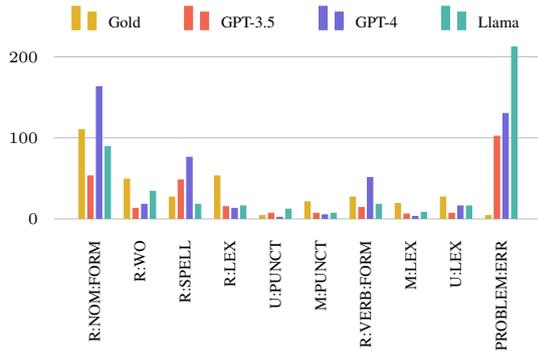
\begin{figure}
\begin{tikzpicture}
	\pgfplotstabletranspose\datatransposed{\datatableErr} 
    \begin{axis}[
        width  = 0.5\textwidth,
        height = 4cm,
        major x tick style = transparent,
        x axis line style = {opacity=0},
        y axis line style = {opacity=0},
        axis y line = left,
        ylabel near ticks,
        ymajorgrids = true,
        ytick style={draw=none},
        xticklabels = {
            R:NOM:FORM,
            R:WO,
            R:SPELL,
            R:LEX,
            U:PUNCT,
            M:PUNCT,
            R:VERB:FORM,
            M:LEX,
            U:LEX,
            PROBLEM:ERR
        },
        xtick = data,
        /pgf/bar width={2pt},  
        ybar=1.25pt,
        xticklabel style = {align=center, font=\tiny, rotate=90},
        yticklabel style = {font=\tiny},
        enlarge x limits=0.1,
        ymin=0, ymax=225,
        legend cell align=left,
        label style={font=\small },
        legend style={
                at={(0.5,1.2)}, anchor=north,
                font=\tiny,
                column sep=0.5ex,
                draw=none,
                legend columns = -1
        },
        legend style={/tikz/every even column/.append style={column sep=0.5cm}}]
    ]
        \addplot[style={tartunlp_yellow, fill=tartunlp_yellow,mark=none}]
             table[y index=1]{\datatransposed};

        \addplot[style={tartunlp_red,fill=tartunlp_red,mark=none}]
             table[y index=2]{\datatransposed};

        \addplot[style={tartunlp_violet,fill=tartunlp_violet,mark=none}]
             table[y index=3]{\datatransposed};            

        \addplot[style={tartunlp_green,fill=tartunlp_green,mark=none}]
             table[y index=4]{\datatransposed};



        \legend{Gold, GPT-3.5, GPT-4, Llama}
    \end{axis}
\end{tikzpicture}
    \caption{Error type count in Estonian based on annotating 100 randomly selected sentences (R - replaced, M - missing, U - unnecessary)}
    \label{fig:human-eval}
\end{figure}

\section{Conclusion}
\label{sct:conclusion}

In conclusion, our research demonstrates the significant potential of Llama-based LLMs in addressing the challenges of GEC for low-resource settings. We have successfully developed state-of-the-art systems for Estonian, Ukrainian, and German by leveraging these models as both correctors and synthetic data generators. We also explore other methods for AEG and show that prompting stronger commercial LLMs is another way of generating high-quality data, and fine-tuning smaller models also has potential when the resources are more limited.

\section{Limitations}
\label{sct:limitations}

Our work focuses on three languages, recognizing that numerous other languages with grammar error correction (GEC) datasets exist outside our study's scope. We selected languages based on recent relevant research activities: Ukrainian due to its recent Shared Task; Estonian, a newly emerging language in GEC research; and German for comparison with a robust 13B model. To comprehensively validate our method, further exploration across additional languages is necessary.

Our objective was not to devise the optimal system exhaustively. Therefore, several avenues remain unexplored, such as varying generation methods, testing different temperatures, and adjusting parameters. Moreover, we capped the generation of synthetic sentences at one million, below the volume utilized in many (though not all) synthetic data studies. Questions about the ideal amount of data needed its dependency on the quality of synthetic and gold examples, remain unanswered.

Furthermore, our study lacks human evaluation of GEC systems, a component for more reliably assessing the real-world efficacy of GEC systems.

\section{Acknowledgements}

This work was partially supported by the Estonian Research Council grant PRG2006 (Language Technology for Low-Resource Finno-Ugric Languages and Dialects) as well as the Institute of the Estonian Language grant LLTAT24472 (Autocorrect for Estonian as a 2nd language for students and teachers). 

We acknowledge the EuroHPC Joint Undertaking for awarding this project access to the EuroHPC supercomputer LUMI, hosted by CSC (Finland) and the LUMI consortium through a EuroHPC Regular Access call.

\bibliography{anthology,custom}
\bibliographystyle{acl_natbib}

\newpage

\appendix

\section{Training details}
\label{sec:appendix-training}
\subsection{Llama-based models}
\label{sec:appendix-training-llama}
The models are trained on 4 AMD MI250x GPUs (each acting as 2 GPUs).

For fine-tuning, we used a learning rate of 5e-6 linearly decayed to 5e-7 (10\%). The learning rate was selected from \{4e-5, 2e-5, 1e-5, 5e-6, 2.5e-6\} based on highest Estonian GEC development set $F_{0.5}$ score. The models were trained for three epochs, although we chose the first epoch since it almost always achieved the highest $F_{0.5}$ score. Table~\ref{tab:llama-hyp-params} provides an overview of the hyperparameters.

For GEC and AEG fine-tuning, sentences are in non-tokenized format or detokenized (for Estonian and German). The crawled data used for AEG is normalized with Moses \citep{koehn-etal-2007-moses} for Estonian and German.

For continued pre-training, we follow the parameters used by Llammas-base (see Table~\ref{tab:llama-cpt-hyp-params}). The training data is packed to fill the whole sequence length.

\begin{table}[!htp]\centering
\begin{tabular}{lrr}\toprule
Parameter &Value \\\midrule
LR &5e-6 \\
LR\textsubscript{final} &5e-7 \\
LR-schedule &linear \\
Epochs &3 \\
Max sequence length &1024 \\
Batch size (total) & 128 \\
Gradient clipping &1.0 \\
Weight decay &0.1 \\
Optimizer &AdamW \\
Precision &bf16 \\
DeepSpeed & Zero Stage 2 \\
\bottomrule
\end{tabular}
\caption{Llama-based GEC model fine-tuning parameters.}\label{tab:llama-hyp-params}

\end{table}

\begin{table}[!htp]\centering
\begin{tabular}{lrr}\toprule
Parameter &Value \\\midrule
LR &2e-5 \\
LR\textsubscript{final} &2e-6 \\
LR-schedule &linear \\
Updates &19080 \\
Max sequence length &1024 \\
Batch size (total) & 256 \\
Gradient clipping &1.0 \\
Weight decay &0.1 \\
Optimizer &AdamW \\
Precision &bf16 \\
DeepSpeed & Zero Stage 2 \\
\bottomrule
\end{tabular}
\caption{Llama continued pre-training parameters.}\label{tab:llama-cpt-hyp-params}

\end{table}

\subsection{NLLB-based models}
\label{sec:appendix-training-nllb}

We follow the training process specified by \citet{luhtaru2024nelb}, including hyperparameters. The training is conducted on an AMD MI250x GPU. We are training the AEG models for 20 epochs and picking the 15th after arbitrary manual evaluation and testing sets on checkpoints 5, 10, 15, and 20. The data for NLLB models is first normalized with Moses script
, and we use the SentencePiece model \cite{kudo-richardson-2018-sentencepiece} for untokenized text. 

\subsection{mT5-based models}
\label{sec:appendix-training-mt5}

To learn to generate errors, we train on reversed human GEC data for three epochs with batch size 32, max sequence length of 128, half-precision training, and a learning rate of 0.0001 without warmup and scheduling. For generation, we use top 50 probabilistic sampling.

\section{Problematic edits}
\label{sec:human-eval}

We further explore the human annotation results discussed in section \ref{results-human-eval}. Table \ref{tab:problematic} displays the percentage of problematic error types out of all errors generated by the model.

\begin{table}[h]
\centering
\begin{tabular}{@{}lccc@{}}
\toprule
Type & Llama & GPT-3.5 & GPT-4 \\ 
\midrule
O & 10.83 & 4.71 & 9.07 \\
HALL & 22.72 & 11.11 & 3.75 \\
SYN & 6.16 & 6.4 & 7.5 \\
INACC & 2.12 & 5.39 & 1.38 \\
UNREC & 3.82 & 6.73 & 3.94 \\
\midrule
Total \% & 45.65 & 34.34 & 25.64 \\ \bottomrule
\end{tabular}
\caption{Percentages of problematic edits.}
\label{tab:problematic}
\end{table}

\section{NLLB correction}
\label{sec:nllb-zs-correction}

The GEC performance of the NLLB model without any synthetic data is in Table \ref{tab:nllb-zs}. The zero-shot results for Estonian and German are significantly higher than for Ukrainian. We notice that the Ukrainian dataset contains characters not present in NLLB vocabulary, like special quotation marks, which the normalization script unifies but appear as errors while testing. In addition, the Ukrainian test set contains far fewer edits, which, especially in a zero-shot scenario, means worse scores because NLLB paraphrases more rigorously \cite{luhtaru2024nelb}.

\begin{table}[h!]
    \centering
    \begin{tabular}{lccc}
    \toprule
       Lang & P & R & F$_{0.5}$ \\
       \midrule
        Estonian & 43.89 & 45.31 & 44.17 \\
        Ukrainian & 8.24 & 31.57 & 9.67 \\
        German & 43.66 & 41.52 & 43.22 \\
    \bottomrule
    \end{tabular}
    \caption{Zero-shot scores of NLLB-200-1.3B-Distilled models on Ukrainian UA-GEC gec+fluency test set.}
    \label{tab:nllb-zs}
\end{table}

\section{Prompts}
\label{sec:appendix-prompts}

We present the prompts used to generate 1) 100,000 sets with GPT-3.5-Turbo and 2) preliminary sets with GPT-4-Turbo in Tables \ref{tab:gpt-prompt-est}, \ref{tab:gpt-prompt-deu}, \ref{tab:gpt-prompt-ukr} for Estonian, German, and Ukrainian respectively.  

\begin{table*}[hp]\centering
\begin{small}
\begin{tabular}{p{0.9\linewidth}}\toprule
Muuda sisendteksti, genereerides sinna vigu, mida võib teha eesti keele õppija. Väljundtekstina tagasta sisendtekst, kuhu oled genereerinud vead. Sisendteksti genereeri õigekirja-, grammatika-, sõnavaliku-, sõnajärje-, kirjavahemärgi- ning stiilivigu. Kui sisendtekstis on vigu, siis ära neid paranda, vaid genereeri vigu juurde. Ülesande kohta on neli näidet:\\
\\
Sisendtekst: \texttt{\{correct\}}\\
Väljundtekst: \texttt{\{incorrect\}}\\
\\
Sisendtekst: \texttt{\{correct\}}\\
Väljundtekst: \texttt{\{incorrect\}}\\
\\
Sisendtekst: \texttt{\{correct\}}\\
Väljundtekst: \texttt{\{incorrect\}}\\
\\
Sisendtekst: \texttt{\{correct\}}\\
Väljundtekst: \texttt{\{incorrect\}}\\
\\
Sisendtekst: \texttt{\{input\}}\\
Väljundtekst:\\
\bottomrule
\end{tabular}
\end{small}
\caption{GPT prompt - Estonian.\label{tab:gpt-prompt-est}}
\end{table*}

\begin{table*}[hp]\centering
\begin{small}
\begin{tabular}{p{0.9\linewidth}}\toprule
Erzeugen Sie im Eingabetext Fehler, wie sie jemand, der Deutsch lernt, machen könnte. Geben Sie als Ausgabetext den Eingabetext zurück, in den Sie Fehler eingefügt haben. Erzeugen Sie Rechtschreib-, Grammatik-, Wortwahl-, Wortreihenfolge-, Zeichensetzungs- und Stilfehler im Eingabetext. Sollten im Eingabetext bereits Fehler vorhanden sein, korrigieren Sie diese nicht, sondern erzeugen Sie zusätzliche Fehler. Es gibt vier Beispiele für die Aufgabe:\\
\\
Eingabetext: \texttt{\{correct\}}\\
Ausgabetext: \texttt{\{incorrect\}}\\
\\
Eingabetext: \texttt{\{correct\}}\\
Ausgabetext: \texttt{\{incorrect\}}\\
\\
Eingabetext: \texttt{\{correct\}}\\
Ausgabetext: \texttt{\{incorrect\}}\\
\\
Eingabetext: \texttt{\{correct\}}\\
Ausgabetext: \texttt{\{incorrect\}}\\
\\
Eingabetext: \texttt{\{input\}}\\
Ausgabetext:\\
\bottomrule
\end{tabular}
\end{small}
\caption{GPT prompt - German.\label{tab:gpt-prompt-deu}}
\end{table*}

\begin{table*}[hp]\centering
\begin{small}
\begin{tabular}{p{0.9\linewidth}}\toprule
\foreignlanguage{ukrainian}{Змініть вхідний текст шляхом генерації в ньому помилок, які міг би зробити учень, що вивчає українську мову. На виході повертайте вхідний текст, у який ви внесли помилки. У вхідному тексті генеруйте помилки правопису, граматики, вибору слів, порядку слів, розділових знаків та стилю. Якщо у вхідному тексті є помилки, то не виправляйте їх, а генеруйте додаткові помилки. Далі наведені чотири приклади до цієї задачі}\\
\\
\foreignlanguage{ukrainian}{Вхідний текст:} \texttt{\{correct\}}\\
\foreignlanguage{ukrainian}{Вихідний текст:} \texttt{\{incorrect\}}\\
\\
\foreignlanguage{ukrainian}{Вхідний текст:} \texttt{\{correct\}}\\
\foreignlanguage{ukrainian}{Вихідний текст:} \texttt{\{incorrect\}}\\
\\
\foreignlanguage{ukrainian}{Вхідний текст:} \texttt{\{correct\}}\\
\foreignlanguage{ukrainian}{Вихідний текст:} \texttt{\{incorrect\}}\\
\\
\foreignlanguage{ukrainian}{Вхідний текст:} \texttt{\{correct\}}\\
\foreignlanguage{ukrainian}{Вихідний текст:} \texttt{\{incorrect\}}\\
\\
\foreignlanguage{ukrainian}{Вхідний текст:} \texttt{\{input\}}\\
\foreignlanguage{ukrainian}{Вихідний текст:} \\
\bottomrule
\end{tabular}
\end{small}
\caption{GPT prompt - Ukrainian. \label{tab:gpt-prompt-ukr}}
\end{table*}

\begin{table*}[h]\centering
\begin{small}
\begin{tabular}{p{0.9\linewidth}}\toprule
\#\#\# Instruction:\\
Reply with a corrected version of the input sentence in \texttt{\{language\}} with all grammatical and spelling errors fixed. If there are no errors, reply with a copy of the original sentence.\\
\\
\#\#\# Input:\\
\texttt{\{input\}}\\
\\
\#\#\# Response:\\
\texttt{\{correction\}}\\
\bottomrule
\end{tabular}
\end{small}
\caption{Llama-based model GEC instruction format loosely based on Alpaca \citep{alpaca}. The instruction is based on \citet{coyne2023analyzing}.}\label{tab:llama-instruction-format}
\end{table*}

\begin{table*}[h]\centering
\begin{small}
\begin{tabular}{p{0.9\linewidth}}\toprule
\#\#\# Instruction:\\
Reply with a grammatically incorrect version of the \texttt{\{language\}} input sentence.\\
\\
\#\#\# Input:\\
\texttt{\{input\}}\\
\\
\#\#\# Response:\\
\texttt{\{correction\}}\\
\bottomrule
\end{tabular}
\end{small}
\caption{Llama-based model AEG instruction format loosely based on Alpaca \citep{alpaca}.}\label{tab:llama-instruction-format-aeg}
\end{table*}

\end{document}